\documentclass[conference]{IEEEtran}

\usepackage{graphicx} 
\usepackage{grffile}
\usepackage{amsmath}
\usepackage{amssymb}
\usepackage{amsfonts}
\usepackage{bm}
\usepackage{accents}
\usepackage{algorithm}
\usepackage{algpseudocode}

\usepackage{url}
\usepackage[colorlinks=true, linkcolor=blue, citecolor=blue, urlcolor=blue]{hyperref}

\DeclareMathOperator*{\argmin}{arg\,min}

\title{Efficient Implementation of Reinforcement Learning over Homomorphic Encryption}

\author{
\IEEEauthorblockN{Jihoon Suh\IEEEauthorrefmark{1}, Takashi Tanaka\IEEEauthorrefmark{1}}
\IEEEauthorblockA{\IEEEauthorrefmark{1}School of Aeronautics and Astronautics, Purdue University, West Lafayette, USA \\
E-mail: suh95@purdue.edu, tanaka16@purdue.edu}
}

\thanks{This manuscript is a preprint version of an original article published in the \textit{Journal of The Society of Instrument and Control Engineers}, vol. 64, no. 4, pp. 223-229, 2025. The original publication is available online at \href{https://doi.org/10.11499/sicejl.64.223}{DOI: 10.11499/sicejl.64.223}}

\begin{document}
\maketitle

\begin{abstract}
We investigate encrypted control policy synthesis over the cloud. While encrypted control implementations have been studied previously, we focus on the less explored paradigm of privacy-preserving control synthesis, which can involve heavier computations ideal for cloud outsourcing. We classify control policy synthesis into model-based, simulator-driven, and data-driven approaches and examine their implementation over fully homomorphic encryption (FHE) for privacy enhancements. A key challenge arises from comparison operations (min or max) in standard reinforcement learning algorithms, which are difficult to execute over encrypted data. This observation motivates our focus on Relative-Entropy-regularized reinforcement learning (RL) problems, which simplifies encrypted evaluation of synthesis algorithms due to their comparison-free structures. We demonstrate how linearly solvable value iteration, path integral control, and Z-learning can be readily implemented over FHE. We conduct a case study of our approach through numerical simulations of encrypted Z-learning in a grid world environment using the CKKS encryption scheme, showing convergence with acceptable approximation error. Our work suggests the potential for secure and efficient cloud-based reinforcement learning.
\end{abstract}


\thanks{\textbf{Note}: This manuscript is a preprint version of an original article published in the \textit{Journal of The Society of Instrument and Control Engineers}, vol. 64, no. 4, pp. 223-229, 2025. The original publication is available online at \href{https://doi.org/10.11499/sicejl.64.223}{DOI: 10.11499/sicejl.64.223}}
\section{Introduction}
Homomorphic Encryption (HE) is a class of cryptographic systems that allow users to perform computational operations on encrypted data without decrypting it. This enables clients to outsource computational tasks to semi-trusted third parties (e.g., external cloud servers) without compromising privacy.

Existing HE systems are categorized into Partial HE (PHE), Leveled HE (LHE), and Full HE (FHE), in increasing order of complexity. PHE schemes typically have a small computational overhead but only support addition or multiplication between encrypted numbers (or ciphertexts). LHE schemes support both addition and multiplication, but these operations are typically allowed only a finite number of times before the data is decrypted. FHE allows additions and multiplications over ciphertexts an infinite number of times, usually employing a technique called \emph{bootstrapping}. Currently, bootstrapping incurs significant computational overhead, which limits the use of FHE in real-time applications. Due to its basic operations being limited to addition and multiplication, other elementary operations such as division and comparisons (including sorting and max-min operators) are not easily implementable. These constraints impose moderate restrictions on the class of algorithms to which they are applicable. Nevertheless, HE has demonstrated promising applications across a wide array of fields, such as medical image processing \cite{Vengadapurvaja17efficient}, medical predictive analysis \cite{Bos14private}, health monitoring \cite{Page15cloud}, secure voting systems \cite{Jabbar17design}, and law enforcement \cite{Bosch14sofir}.

\begin{figure*}[!t]
\centering
\includegraphics[width=0.96\textwidth]{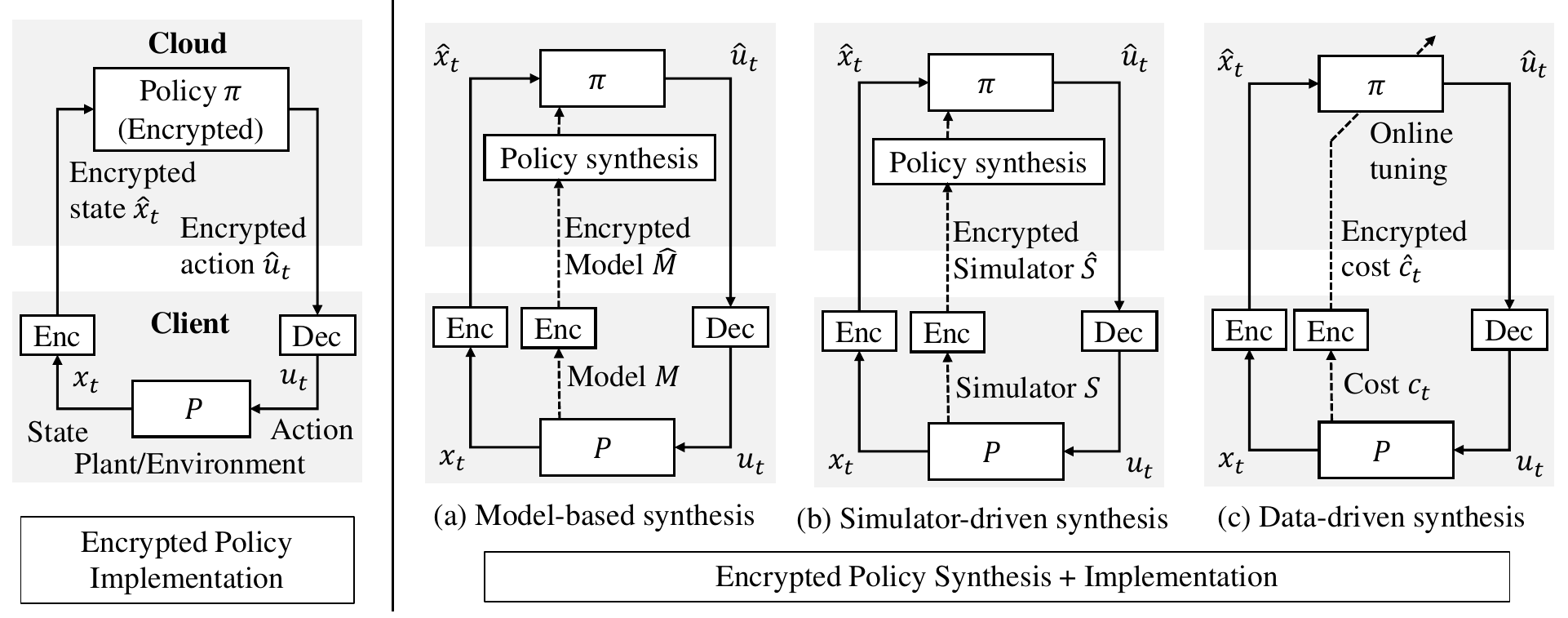}
\caption{Encrypted policy implementation (left) and encrypted policy synthesis (right).}
\label{fig:1}
\end{figure*}

HE for secure cloud-based control was conceptualized in the mid-2010s \cite{Kogiso15cyberencrypting,Kim16encrypting,Farokhi16secure}. The primary focus of early literature on encrypted control was on cloud-based \emph{implementations} of control policies. As shown in Fig.~\ref{fig:1} (left), the role of the cloud is simply to map the encrypted state to the encrypted action based on the already synthesized policy $\pi$.
Since real-time performance is a stringent requirement in control systems, PHE or LHE are typically the main choices in this paradigm.

In this article, we focus on cloud-based policy \emph{synthesis} for control systems. 
Although control policy synthesis over HE has been less explored in the literature, we believe that studying this paradigm is beneficial, as policy synthesis often involves heavy computations that should ideally be performed on the cloud. 
Since the synthesis process requires knowledge of the client's environment, the entire synthesis algorithm must be encrypted to ensure privacy. 
Policy synthesis usually has less stringent real-time requirements than policy implementation, as it can be executed offline. Hence, we assume FHE, which supports unlimited additions and multiplications of encrypted data.

For systematic analysis, we classify control policy synthesis into model-based, simulator-driven, and data-driven approaches. 
In this article, we consider how these methods can be executed homomorphically over ciphertexts, as illustrated in Fig.~\ref{fig:1} (right). 
As representative algorithms from these three categories, we describe three standard reinforcement learning (RL) algorithms in Section~\ref{sec:RL}: Value Iteration, Monte Carlo methods, and Q-learning. A key challenge in implementing these algorithms over FHE arises from their recursive formulas containing comparison operations (e.g., min or max), which are difficult to execute over encrypted data.

This observation motivates us to pay special attention to what we call the Relative Entropy (RE)-regularized RL problems in Section~\ref{sec:RE_RL}. 
In the literature, this class of RL problems has been known under various names, such as \emph{linearly solvable MDPs} \cite{Todorov06linearly} \emph{path integral control} \cite{Kappen05path} \emph{KL control} \cite{Theodorou12relative} and \emph{Z-learning} \cite{Todorov09efficient}. We show that model-based, simulator-driven, and data-driven synthesis in the realm of RE-regularized RL can be drastically simplified compared to their generic counterparts in Section~\ref{sec:RL}. Importantly, their recursive formulas are free from comparison operators and can be readily implemented over FHE.

\section{Generic RL Methods and Their Homomorphic Implementations}
\label{sec:RL}
We begin by reviewing generic (tabular) RL methods \cite{Sutton18reinforcement} and identifying technical difficulties when they were to be implemented on HE.

\subsection{Problem setup}
Consider a time-invariant Markov Decision Process (MDP) over discrete time steps $t=0,1,2,\cdots$ defined by a tuple $(\mathcal{X}, \mathcal{U}, P, C)$, where $\mathcal{X}$ is a finite state space, $\mathcal{U}$ is a finite action space, $P(x_{t+1}|x_t, u_t)$ is the transition probability, and $C(x_t, u_t)$ is the cost.

A \emph{policy} $\pi=(\pi_0,\pi_1,\pi_2,\cdots)$ is a sequence of conditional probability distributions, where each $\pi_t(u_t|x_{0:t}, u_{0:t-1})$ specifies the probability of selecting action $u_t\in\mathcal{U}$ given the history of states and actions $(x_{0:t}, u_{0:t-1})$.
The \emph{value} of a policy $\pi$ is a function $V^\pi:\mathcal{X}\rightarrow \mathbb{R}$ defined by
\begin{equation}
\label{eq:value_pi}
V^\pi(x)=\mathbb{E}^\pi\left[\sum_{t=0}^\infty \gamma^t C(x_t, u_t) \Big| x_0=x\right]
\end{equation}
where $0\leq \gamma < 1$ is a predefined discount factor.
We say that a policy $\pi$ is optimal if it minimizes the value \eqref{eq:value_pi} of each state $x\in\mathcal{X}$ simultaneously. An optimal policy is denoted by $\pi^*$, and the goal of the policy synthesis is to compute $\pi^*$ from an explicit model of the MDP (model-based RL), simulators of the MDP (simulator-driven RL), or operational data of the MDP (data-driven RL).

The existence of time-invariant, Markov and deterministic optimal policy is well-known \cite{Puterman14markov}. This allows us to restrict our attention to the class of state feedback policies of the form $u_t=\pi(x_t)$ without loss of performance.
The value of the optimal policy $\pi^*$, denoted as $V^*:\mathcal{X}\rightarrow \mathbb{R}$, is called the \emph{optimal value function}.

\subsection{Model-based RL: Value iteration}
\label{sec:VI_basic}
By Bellman's optimality principle, the value function satisfies 
\begin{equation}
\label{eq:bellman}
V^*(x)=(TV^*)(x) \; \forall x \in \mathcal{X}
\end{equation}
where the operator $T$ is defined by
\begin{equation*}
(TV)(x)=\min_{u\in\mathcal{U}} \left[C(x,u)+\gamma \sum_{x'\in\mathcal{X}} P(x'|x,u)V(x')\right].
\end{equation*}
The operator $T$ is known to be contractive in that $\|TV-T\bar{V}\|_\infty \leq \gamma \|V-\bar{V}\|$ for any vectors $V$ and $\bar{V}$.
This implies that the value function can be computed by repeating the \emph{value iteration}
\begin{equation}
\label{eq:VI}
V^{k+1}(x)=TV^k(x) \; \forall x\in \mathcal{X}
\end{equation}
until convergence. Convergence occurs after a finite number of iterations if $\mathcal{X}$ and $\mathcal{U}$ are finite. The optimal policy can be obtained from $V^*$ as
\begin{equation}
\label{eq:pi_v}
\pi^*(x)=\argmin_{u\in\mathcal{U}} \left[C(x,u)+\gamma \sum_{x'\in\mathcal{X}} P(x'|x,u)V(x')\right].
\end{equation}

\subsection{Simulator-driven RL: Monte-Carlo method}
\label{sec:MC_basic}
Value iteration is often impractical in many real-world scenarios because the exact transition probability $P(x_{t+1}|x_t, u_t)$ is frequently unavailable or difficult to obtain.
However, even in such cases, it is often easy to simulate a full episode $x_0, u_0, x_1, u_1, \cdots$ under a given policy $\pi$ (e.g., blackjack \cite{Sutton18reinforcement}). Monte Carlo RL comprises a class of algorithms that compute $\pi^*$ from a large number of simulated episodes.

Note that the optimal policy $\pi^*$ may not be readily computable from the value function $V^*$ using the formula \eqref{eq:pi_v} if the model $P(x_{t+1}|x_t, u_t)$ is unavailable. Therefore, instead of $V^*$, Monte Carlo methods attempt to compute the Q-function:
\begin{equation}
Q^*(x,u)=C(x,u)+\gamma \sum_{x'\in\mathcal{X}}P(x'|x,u)V^*(x)
\end{equation}
which quantifies the optimal value of each state-action pair $(x,u)$. Algorithm~\ref{algo:MCES} is called the Monte-Carlo ES (exploring start) \cite{Sutton18reinforcement}, one of the simplest simulator-driven RL methods.
This algorithm belongs to the general class of policy iteration methods, with notable features that the initial condition explores all possible state-action pairs, and that the Q-function corresponding to the current policy is directly estimated from Monte Carlo simulations. 
Once the estimate of the Q-function converges, the desired policy $\pi^*$ can be readily recovered by the operation in line 9. 
\begin{algorithm}[t]
\caption{Monte-Carlo ES}\label{algo:MCES}
\textbf{Input} $L\in\mathbb{N}$, $N\in\mathbb{N}$
\begin{algorithmic}[1]
  \State Initialize $\pi(x)$ and $Q(x,u)$ for all $x\in \mathcal{X}$ and $u\in \mathcal{U}$.
  \For{$\ell = 0, 1, \cdots, L-1$}
  \For{each state-action pair $(x_0, u_0)\in \mathcal{X}\times \mathcal{U}$}
  \State Generate episodes $(x_0, u_0, x_1^i, u_0^i, x_2^i, u_2^i, \cdots)$ for $i=1, \cdots, N$ following $\pi$, all starting from $(x_0, u_0)$.
  \State Observe costs $G^i=\sum_{t=0}^\infty \gamma^t C(x_t^i, u_t^i)$ $i=1, \cdots, N$.
  \State $Q(x_0, u_0)=\frac{1}{N}\sum_{i=0}^N G^i$
  \EndFor
  \For{$x\in\mathcal{X}$}
  \State $\pi(x)=\argmin_u Q(x,u)$
  \EndFor
  \EndFor
\end{algorithmic}
\textbf{Output} $\pi$
\end{algorithm}

The conditions under which Monte Carlo ES converges to $Q^*$ have been studied by many authors \cite{Tsitsiklis02}. Note that Algorithm~\ref{algo:MCES} is only conceptual in that it requires infinitely long simulated episodes to compute the empirical costs.
As a practical alternative, it has been suggested \cite{Tsitsiklis02} to stop the simulation with probability $1-\gamma$ at every time step (which ensures that each simulation is terminated after a finite number of time steps with probability one) and to use undiscounted summation in line 5.  

\subsection{Data-driven RL: Temporal difference method}
\label{sec:TD_basic}
Temporal difference (TD) learning is similar to Monte Carlo methods in that it does not require explicit knowledge of the underlying MDP.
However, unlike Monte Carlo methods, TD learning allows the Q-function to be updated at every time step without waiting until the end of the simulation of the full episode. 
This is a convenient feature for data-driven synthesis, where the policy is updated as soon as new data (from the environment) arrives. The TD counterpart of Algorithm~\ref{algo:MCES} is $Q$-Learning, which is summarized as Algorithm~\ref{algo:Q}.

\begin{algorithm}[t]
\caption{Q-learning}\label{algo:Q}
\textbf{Input} Small $\epsilon>0$ and step sizes $\alpha_t(x,u)$ for each $t=0,1,\cdots$ and $(x,u)\in\mathcal{X}\times\mathcal{U}$.
\begin{algorithmic}[1]
  \State Initialize $Q(x,u)$ for all $x\in \mathcal{X}$ and $u\in \mathcal{U}$.
  \State Initialize $x_0\in\mathcal{X}$.
  \For{$t= 0, 1, 2, \cdots$}
  \State Choose $u_t=\pi_t^\epsilon(x_t)$ ($\epsilon$-greedy policy).
  \State Take action $u_t$, and observe $C_t=C(x_t, u_t)$ and $x_{t+1}$.
  \State Set $Q_{t+1}(x_t,u_t)=(1-\alpha_t(x_t,u_t))Q_t(x_t,u_t)+\alpha_t(x_t,u_t)\left[C_t+\gamma \min_{u} Q_t(x_{t+1},u)\right]$.
  \State Set $Q_{t+1}(x,u)=Q_t(x,u)$ for all $(x,u)\neq (x_t, u_t)$.
  \EndFor
\end{algorithmic}
\end{algorithm}

It is known that $Q_t$ converges to $Q^*$ under the Robbins-Monroe condition
\begin{equation*}
\sum_{t=0}^\infty \alpha_t(x,u)=\infty,  \sum_{t=0}^\infty \alpha_t^2(x,u)<\infty \; \forall (x,u)\in\mathcal{X}\times \mathcal{U}.  
\end{equation*}
Consequently, the estimate of the optimal policy
\begin{equation*}
\pi_t(x)=\argmin_{u\in\mathcal{U}} Q_t(x,u)
\end{equation*}
converges to $\pi^*$. Q-learning is considered an off-policy RL algorithm because the target policy $\pi_t$ differs from the behavior policy used in the learning process. For example, the $\epsilon$-greedy policy:
\begin{equation*}
\pi_t^\epsilon(x)=\begin{cases} \pi_t(x) & \text{with probability } 1-\epsilon \\
u\sim \text{Unif}(\mathcal{U}) & \text{with probability } \epsilon
\end{cases}
\end{equation*} 
is often chosen as a behavior policy for action selection (line 4 of Algorithm~\ref{algo:Q}) to facilitate sufficient exploration.

\subsection{Homomorphic implementation}
We now discuss the feasibility of implementing the RL algorithms described so far using FHE. Specifically, we wish to implement value iteration \eqref{eq:VI}, Monte Carlo ES (Algorithm~\ref{algo:MCES}), and Q-Learning (Algorithm~\ref{algo:Q}) within the architecture shown in Fig.~\ref{fig:1} (right). Ideally, these iterative algorithms would be executable entirely within the ciphertext domain, without the need to decrypt intermediate results. However, all three algorithms share a fundamental limitation, as they all require ``min" operators in their main iterations. 
Since comparison operations are usually not supported by FHE, it is often necessary to employ \emph{ad hoc} solutions --- such as requiring the client to decrypt intermediate data for comparison in plaintext \cite{Suh21sarsa,Suh21encrypted} --- which clearly diminishes the utility of cloud computing. 

Motivated by these limitations, we next focus on a special class of control problems that enables the implementations of model-based, simulator-driven, and data-driven policy syntheses exclusively over FHE.

\section{Relative-Entropy-Regularized Reinforcement Learning}
\label{sec:RE_RL}
In this section, we focus on what we call the Relative-Entropy (RE)-regularized RL problems.
This class of policy synthesis problems is also known as \emph{linearly solvable MDPs} \cite{Todorov09efficient} or \emph{KL control problems} \cite{Theodorou12relative} and is also related to the paradigm of \emph{path integral control} \cite{Kappen05path}
RE-regularized RL imposes certain structural assumptions on the problem setup. 
However, these assumptions allow for a drastic simplification of the learning algorithms.
Importantly, the simplification includes the removal of the comparison operators from the learning algorithms, which makes their homomorphic implementations significantly easier.

\subsection{Problem setup}
As before, we consider an MDP $(\mathcal{X}, \mathcal{U}, P, C)$ with a finite state space $\mathcal{X}$ and a finite action space $\mathcal{U}$. 
To ensure that each episode under the optimal policy terminates in a finite number of time steps, we also assume that there exists an absorbing state $x_{\text{abs}}\in\mathcal{X}$. 
We initialize the state $x_0\in \mathcal{X}\setminus \{x_{\text{abs}}\}$ and let the random time $T_f$ be the smallest time step $t$ such that $x_t = x_{\text{abs}}$.
We are interested in computing a policy $\pi$ that minimizes the value 
\begin{align}
&V^\pi(x)= \nonumber\\
&\mathbb{E}^\pi \left[\sum_{t=0}^{T_f-1}\left\{C(x_t, u_t)+\lambda \log\frac{\pi(u_t|x_t)}{b(u_t|x_t)}\right\}\Big| x_0=x \right] \label{eq:KLD_RL}
\end{align}
simultaneously for all $x\in\mathcal{X}\setminus \{x_\text{abs}\}$. We impose $V^\pi(x_\text{abs})=0$ for all $\pi$, assuming that the absorbing state incurs no cost.
In \eqref{eq:KLD_RL}, $\mathbb{E}\log\frac{\pi}{b}=D(\pi\|b)$ is relative entropy (RE) and $\lambda$ is a positive regularizer. The RE term prevents $\pi$ from deviating too much from a fixed policy $b$, which is given as part of the problem setup. 
The policy $b$ is called \emph{passive dynamics} or \emph{behavior policy} in the literature and will be used to explore the environment during the learning process. 

We also make the following assumptions:
\begin{itemize}
\item[(i)] The transition $P(x_{t+1}|x_t, u_t)$ is deterministic. That is, there exists a function $F$ such that $x_{t+1}=F(x_t, u_t)$.
\item[(ii)] The cost functional \eqref{eq:KLD_RL} is undiscounted.
\end{itemize}
Both of these assumptions are critical for the simplifications below.

\subsection{Model-based RL: Linearly solvable value iteration}
As in Section~\ref{sec:VI_basic}, let us begin with model-based synthesis, assuming that explicit models of $F$, $C$, and $b$ are available. By Bellman's optimality principle, the value function satisfies 
\begin{align}
&V^*(x_t)= \nonumber \\
&\min_{\pi(u_t|x_t)}\sum_{u_t}\pi(u_t|x_t) \left\{\rho(x_t, u_t)+\lambda \log\frac{\pi(u_t|x_t)}{b(u_t|x_t)}\right\}\label{eq:lLMDP_Vstar}
\end{align}
where $\rho(x_t, u_t)=C(x_t, u_t)+V^*(F(x_t,u_t))$.
Unlike the generic value function in Section~\ref{sec:VI_basic}, this equation can be further simplified by noticing that \eqref{eq:lLMDP_Vstar} is the \emph{free energy minimization} problem \cite{Theodorou12relative}. An analytical form of the optimal policy is available and is given as the Boltzmann distribution:
\begin{equation}
\label{eq:Boltzmann}
\pi^*(u_t|x_t)=\frac{b(u_t|x_t)\exp\left\{-\rho(x_t,u_t)/\lambda\right\}}{\sum_{u'_t}b(u'_t|x_t)\exp\left\{-\rho(x_t,u'_t)/\lambda\right\}}.
\end{equation}
Substituting \eqref{eq:Boltzmann} back to \eqref{eq:lLMDP_Vstar}, one can eliminate the ``min" operator, and \eqref{eq:lLMDP_Vstar} can be simplified to
\begin{align*}
&V^*(x_t)=  \\
&-\lambda \log \sum_{u_t}b(u_t|x_t)\exp\left\{-\frac{C(x_t,u_t)}{\lambda}-\frac{V^*(x_{t+1})}{\lambda}\right\}.
\end{align*}
Introducing the \emph{desirability function} $Z^*(x):=\exp\left(-V^*(x)/\lambda\right)$, this equation can be further simplified to
\begin{equation}
\label{eq:Z_linear}
Z^*(x_t)=\sum_{u_t}b(u_t|x_t)\exp\left(-\frac{C(x_t, u_t)}{\lambda}\right)Z^*(x_{t+1})
\end{equation}
for $x_t\in\mathcal{X}\setminus \{x_\text{abs}\}$ and $Z^*(x_\text{abs})=\exp\left(-V^*(x_\text{abs})/\lambda\right)=1$. Remarkably, this is a linear equation of $Z^*$. To make this point explicit, let us index the state space as $\mathcal{X}=\{1, 2, \cdots, S, x_\text{abs}\}$, and let 
\[
\bar{Z}^*=\begin{bmatrix}
Z^*(1) & Z^*(2) & \cdots & Z^*(S)
\end{bmatrix}^\top \in\mathbb{R}^S
\] 
be the vector of desirability values. (We have omitted $Z^*(x_\text{abs})=1$ from the list). Equation \eqref{eq:Z_linear} implies that $\bar{Z}^*$ satisfies
\[
\begin{bmatrix}
Z^*(1) \\ \vdots \\ Z^*(S)
\end{bmatrix}=
\underbrace{\begin{bmatrix}
a_{11} & \cdots & a_{1S} \\ \vdots & &\vdots \\ a_{S1} & \cdots & a_{SS}
\end{bmatrix}}_{A}
\begin{bmatrix}
Z^*(1) \\ \vdots \\ Z^*(S)
\end{bmatrix}
+\underbrace{\begin{bmatrix}
w_1 \\ \vdots \\ w_S
\end{bmatrix}}_{w}.
\]
Entries of $A$ and $w$ are
\begin{align*}
a_{ij}&=\sum\nolimits_{u\in\mathcal{U}_{i\rightarrow j}} b(u|i)\exp\left(-C(i,u)/\lambda \right) \\
w_i&=\sum\nolimits_{u\in\mathcal{U}_{i\rightarrow x_\text{abs}}} b(u|i)\exp\left(-C(i,u)/\lambda\right)
\end{align*}
where
$\mathcal{U}_{i\rightarrow j}=\left\{u\in\mathcal{U} \big| F(i,u)=j\right\}$.
If the matrix $A$ is contractive (it follows from the Perron-Frobenius theorem that $A$ is contractive if $C(x,u)>0$ for all $x\in \mathcal{X}\setminus \{x_\text{abs}\}$ and $u\in\mathcal{U}$), a simple recursion
\begin{equation}
\label{eq:VI_Z}
\bar{Z}_{k+1}=A\bar{Z}_k+w, \; \bar{Z}_0>0 \text{ (arbitrary)}
\end{equation}
achieves convergence $\bar{Z}_k\rightarrow \bar{Z}^*$ as $k\rightarrow \infty$. This implies that, unlike the generic value iteration \eqref{eq:VI} (which involved a ``min" operator), the value iteration \eqref{eq:VI_Z} only requires additions and multiplications.

\subsection{Simulator-driven RL: Path integral control}
Next, let us consider a scenario where the models $F$, $C$, and $b$ are not available but episodes $x_0, u_0, x_1, u_1, \cdots, x_{T_f}$ under the behavioral policy $b$ are easy to simulate.
As in Section~\ref{sec:MC_basic}, Monte Carlo methods are useful in such scenarios.

Equation \eqref{eq:Z_linear} implies that the desirability function at time step $t$ can be expressed in terms of the desirability function at time step $t+1$. Since the same argument holds between time steps $t+1$ and $t+2$, $Z^*(x_t)$ can also be expressed as
\begin{align*}
&Z^*(x_t)=\sum_{u_t}b(u_t|x_t)\exp\left(-\frac{C(x_t, u_t)}{\lambda}\right) \\
&\times\sum_{u_{t+1}}b(u_{t+1}|x_{t+1})\exp\left(-\frac{C(x_{t+1}, u_{t+1})}{\lambda}\right)
Z^*(x_{t+2})
\end{align*}
where $x_{t+1}=F(x_t, u_t)$ and $x_{t+2}=F(x_{t+1}, u_{t+1})$. By repeating this argument from $t=0$ till the end of the episode, we obtain
\begin{align}
Z^*(x_0)&=\sum_{u_{0:T_f-1}}\prod_{t=0}^{T_f-1}b(u_t|x_t)\exp\left\{-\frac{1}{\lambda}\sum_{t=0}^{T_f-1}C(x_t,u_t)\right\} \nonumber \\
&=\mathbb{E}^b \exp\left\{-\frac{1}{\lambda}\sum_{t=0}^{T_f-1}C(x_t, u_t)\right\} 
\label{eq:MC_Zstar}
\end{align}
where $x_{t+1}=F(x_t,u_t)$ for $t=0, 1, \cdots, T_f-1$.

The result \eqref{eq:MC_Zstar} indicates that $Z^*(x_0)$ can be computed by Monte Carlo simulations of full episodes. Specifically, for $i=1, 2, \cdots, N$, let 
\[
x_0, u_0^i, x_1^i, u_1^i, \cdots, x_{T_f-1}^i, u_{T_f-1}^i
\]
be $N$ independently simulated episodes starting from $x_0$ under the behavioral policy $b$.
By the law of large numbers, we expect that the empirical mean converges to $Z^*(x_0)$ quadratically as $N$ tends to infinity:

\begin{equation}
\label{eq:z_mc}
Z^*(x_0)\approx \frac{1}{N}\sum_{i=1}^N \exp\left\{-\frac{1}{\lambda}\sum_{t=0}^{T_f-1}C(x_t^i,u_t^i)\right\}.
\end{equation}
Consequently, the value function $V^*(x)$ of the RE-regularized RL problem \eqref{eq:KLD_RL} can also be computed by Monte Carlo simulations as follows:
\begin{equation}
\label{eq:Vstar_MC}
V^*(x_0)\approx -\lambda \log \left[\frac{1}{N}\sum_{i=1}^N \exp\left\{-\frac{1}{\lambda}\sum_{t=0}^{T_f-1}C(x_t^i,u_t^i)\right\}\right].
\end{equation}
This formulation highlights a key insight: the value function can be estimated directly from simulated path costs (path integrals) generated under the behavior policy. This fundamental idea underpins path integral control algorithms \cite{Kappen05path}.

A major distinction exists between \eqref{eq:Vstar_MC} and the Monte Carlo method (Algorithm~\ref{algo:MCES}) employed in generic RL. As evident from Algorithm~\ref{algo:MCES}, Monte-Carlo-based RL in the generic setup requires a simultaneous search for both the optimal Q-function $Q^{*}$ and the policy $\pi^{*}$, which is achieved via policy iteration.
In contrast, the approach in \eqref{eq:Vstar_MC} circumvents the need for explicit knowledge of the optimal policy $\pi^*$. Since episodes can be generated directly using the known behavior policy $b$, the value function in \eqref{eq:Vstar_MC} can be computed from a single batch of Monte Carlo simulations. This simplifies the overall algorithm substantially. Additionally, \emph{importance sampling} \cite{Kappen05path, Sutton18reinforcement} can be employed to estimate \eqref{eq:MC_Zstar} using samples generated by an alternative behavior policy $b' (\neq b)$.

\subsection{Data-driven RL: Z-learning} \label{data_z}
In Section~\ref{sec:TD_basic}, we applied the concept of TD learning to the Monte Carlo method (Algorithm~\ref{algo:MCES}) and obtained $Q$-learning (Algorithm~\ref{algo:Q}).
In a similar vein, we can apply TD learning to the Monte Carlo method \eqref{eq:Vstar_MC} and obtain the so-called Z-learning algorithm \cite{Todorov09efficient}.

Like Q-learning, Z-learning allows the estimate $\hat{Z}$ of the desirability function to be updated using data (physical or simulated) that become available in each time step.
Let $x_t, u_t, x_{t+1}$ be the transition generated by the behavioral policy $b$, and let $C_t=C(x_t, u_t)$ be the observed cost. 
The estimate $\hat{Z}$ is updated online as
\begin{equation}
\label{eq:zlearning}
\hat{Z}_{t+1}(x_t)=(1-\alpha_t)\hat{Z}_t(x_t)+\alpha_t \exp\left(-\frac{C_t}{\lambda}\right)\hat{Z}_t(x_{t+1})
\end{equation}
where $\alpha_t$ is a diminishing learning rate.

\subsection{Homomorphic implementation}
It is worth noting that the main iterations of the value iteration \eqref{eq:VI_Z}, Monte Carlo estimates \eqref{eq:z_mc}, and TD learning \eqref{eq:zlearning} in the RE-regularized RL regime do not involve ``min" operators. This is in stark contrast to the generic counterparts considered in Section~\ref{sec:RL}.
The absence of ``min" operators is a significant advantage for their implementations over FHE. While formulas \eqref{eq:VI_Z}, \eqref{eq:z_mc}, and \eqref{eq:zlearning} may involve exponential operations, our experiments show that polynomial expansions often suffice to approximate these operations.

\section{Numerical Study}
In this section, we present the numerical simulation of an encrypted Z-learning (see Section \ref{data_z}) over HE in a client-server architecture. We consider a grid world maze problem with the following setup:

\begin{itemize}
    \item State space $\mathcal{X}$: A $9 \times 9$ discrete grid world is configured, and each grid that is neither a trap state (marked by the letter \lq{T}\rq) nor a goal state (terminal state; marked by the letter \lq{G}\rq) is a valid state $x \in \mathcal{X}$.
    \item Action space $\mathcal{U}$: Nine (or fewer) discrete actions are allowed at each state. They include four cardinal directions, four diagonal directions, and one \textit{Stay} action. Note that an action that leads to a trap state is prohibited.
    \item Cost function $C(x_t, u_t)$: Each transition incurs a fixed running cost, except for the transition to the goal state, which terminates the episode.
\end{itemize}

At the start of the episode, the agent is spawned at a random coordinate in the maze. The agent runs a passive dynamics as the behavioral policy, with uniform random sampling from $\mathcal{U}$, such that $b(u_t|x_t) = \frac{1}{|\mathcal{U}|}, \: \forall u_t \in \mathcal{U}$, until it terminates. The episode terminates when the agent takes the maximum steps allowed per episode or reaches a terminal state. The following hyperparameters were used:
\begin{itemize}
    \item Maximum steps per episode: $200$
    \item Learning rate: $\alpha_t(x_t, k) = \frac{1000}{1000 + n(x_t, k)}$, where $n(x_t, k)$ is the number of visits to the state $x_t$ at the step $k$.
    \item Number of episodes: $5000$
    \item Regularization constant: $\lambda = 0.15$ 
\end{itemize}

We employed the CKKS \cite{CKKS} encryption scheme, a leveled HE scheme designed to support approximate arithmetic over complex numbers. Under the leveled HE scheme, we have a limited multiplicative depth, and thus we approximate the Z-learning update rule via Taylor series approximations. The simulation was conducted using Microsoft SEAL library \cite{SEAL22} under the polynomial modulus degree $N=2^{14}$, the coefficient modulus $Q$ as a chain of prime moduli with bit sizes $\{60, 30, 30, 30, 30, 60\}$, and a scaling factor $\Delta = 2^{40}$ to evaluate multiple ciphertext arithmetic in sufficient precision.

\begin{figure}[t]
\centering
\includegraphics[width=\columnwidth]{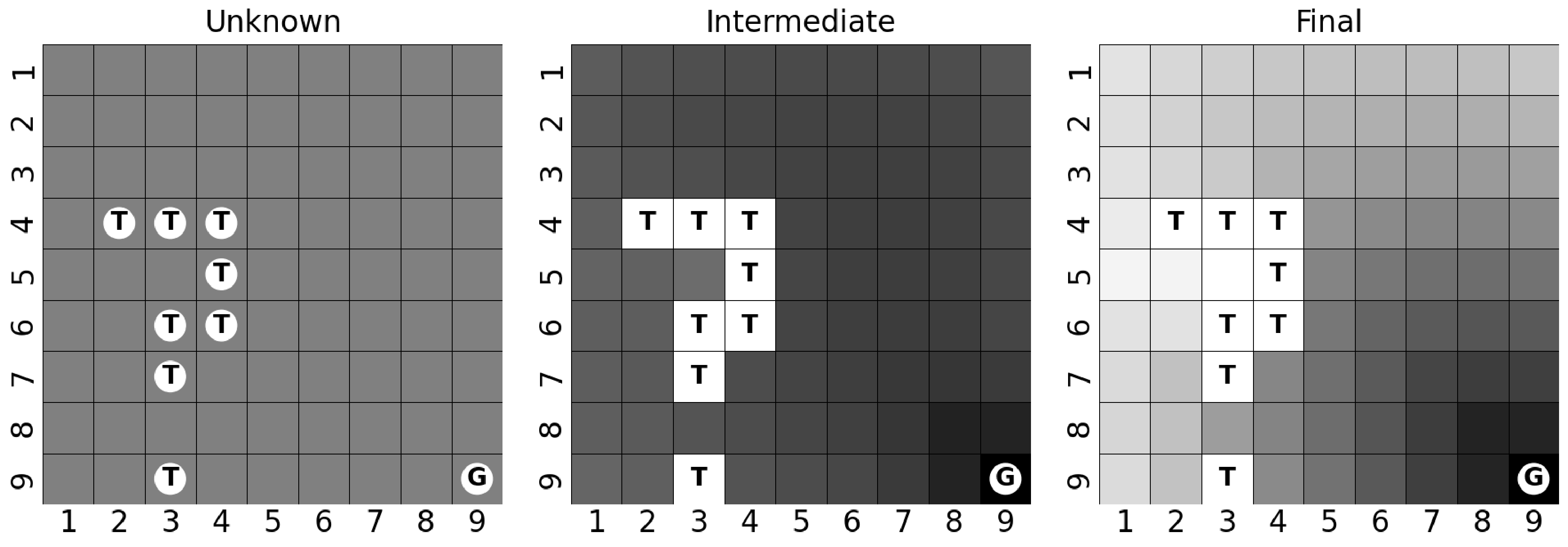}
\caption{Grid World Value Progression (Z-Learning).}
\label{fig:2}
\end{figure}

\begin{figure}[t]
\centering
\includegraphics[width=\columnwidth]{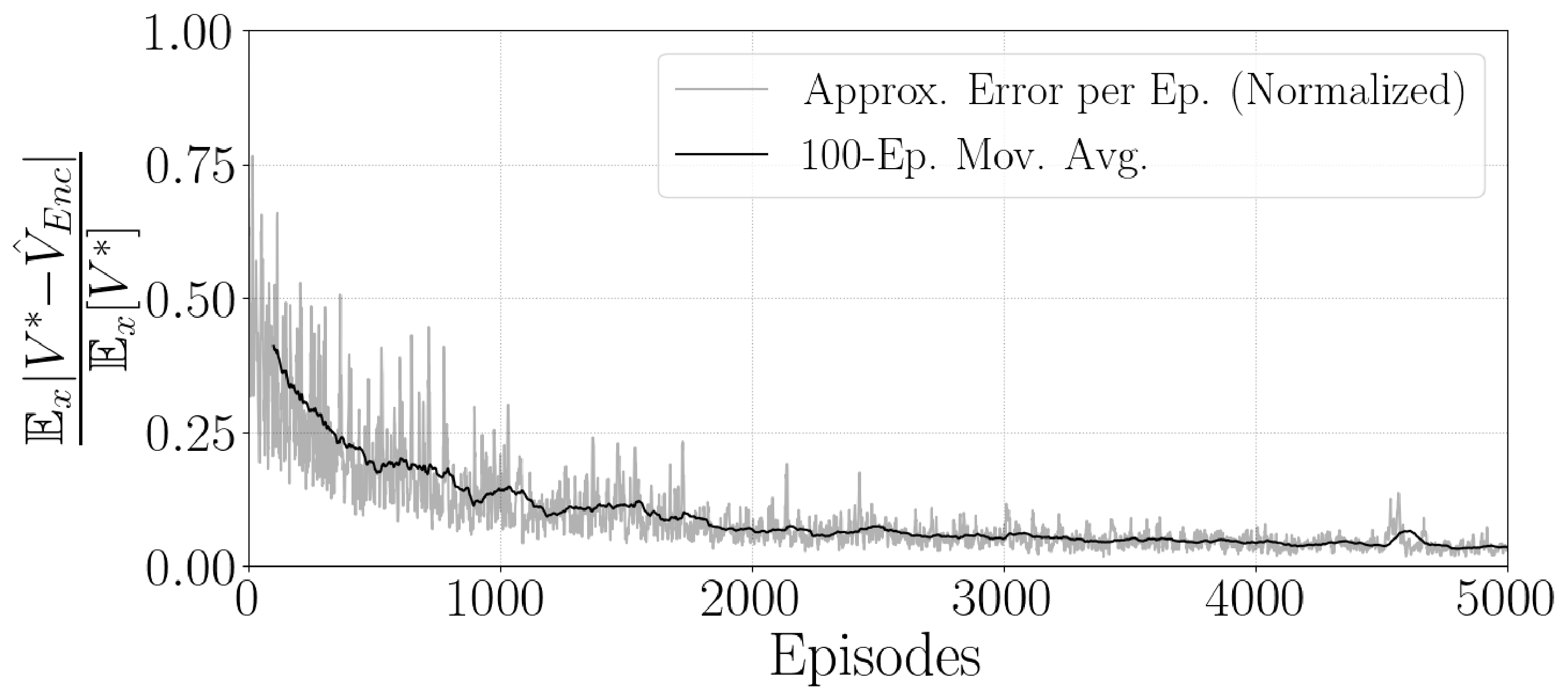}
\caption{Approximation Error Normalized.}
\label{fig:3}
\end{figure}

In Fig~\ref{fig:2}, we visualize the progression of value function estimates during Z-learning. Furthermore, the optimal value function was obtained using the value iteration \eqref{eq:VI_Z}. The optimal value was used to compute the normalized approximation error $\frac{\mathbb{E}_{x}|V^{*} - V_{\text{Enc}}|}{\mathbb{E}_{x}[V^{*}]}$, where $V_{\text{Enc}}$ is the value estimated by the encrypted implementation of \eqref{eq:zlearning}. The normalized approximation error per episode is shown in Fig~\ref{fig:3}.

\section{Summary and Future Work}
This article has provided an overview of RL methods for RE-regularized policy synthesis problems and compared them to generic tabular RL methods in finite MDPs. We argued that the algorithmic simplicity of RE-regularized methods makes them particularly well-suited for FHE implementation, suggesting the potential for secure and efficient cloud-based applications. Our future work will include the experimental validation of this assertion.

\bibliographystyle{IEEEtran}
\bibliography{encRL_sice}

\begin{IEEEbiography}[{\includegraphics[width=1in,height=1.25in,clip,keepaspectratio]{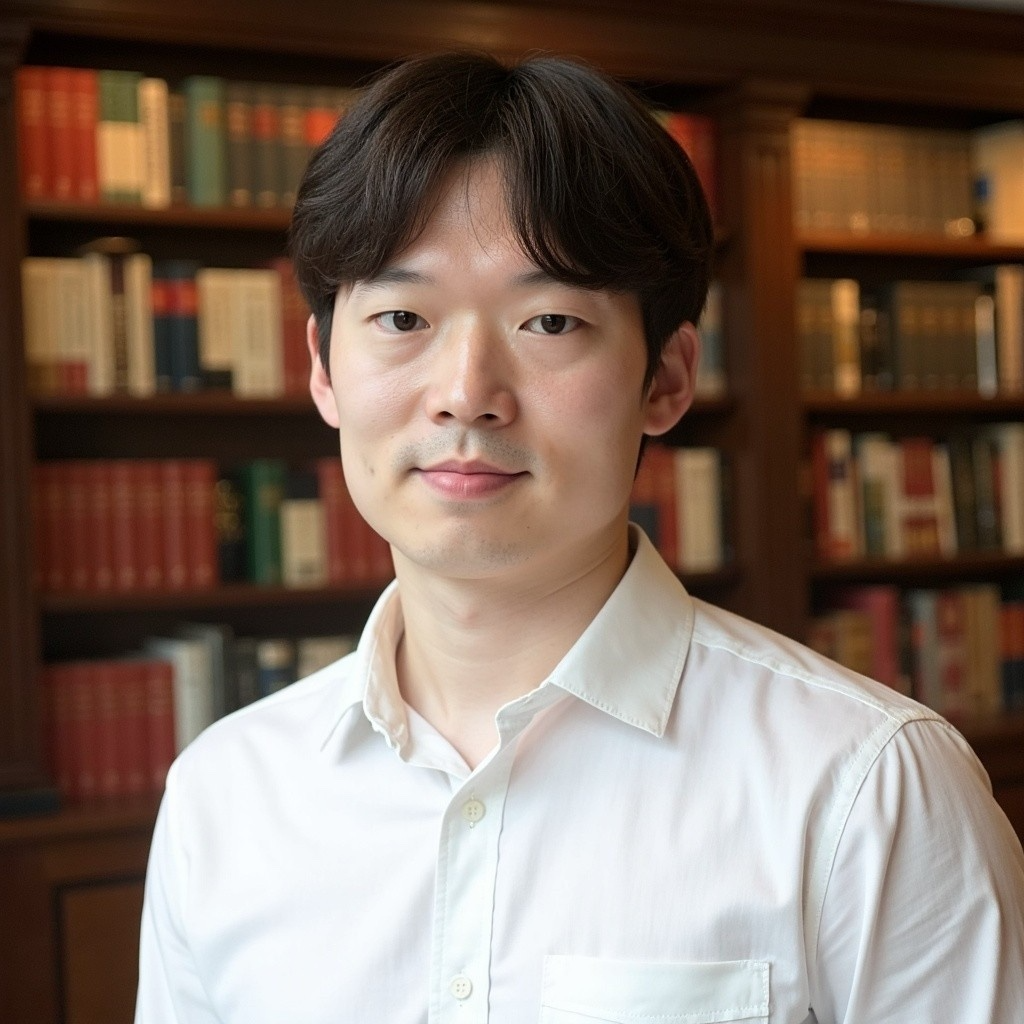}}]{Jihoon Suh}
Jihoon Suh is a Ph.D. student in the School of Aeronautics and Astronautics at Purdue University. He received his B.S. (2018) and M.S. (2020) degrees from the University of Texas at Austin. His research focuses on the development of encrypted control and its applications.
\end{IEEEbiography}

\begin{IEEEbiography}[{\includegraphics[width=1in,height=1.25in,clip,keepaspectratio]{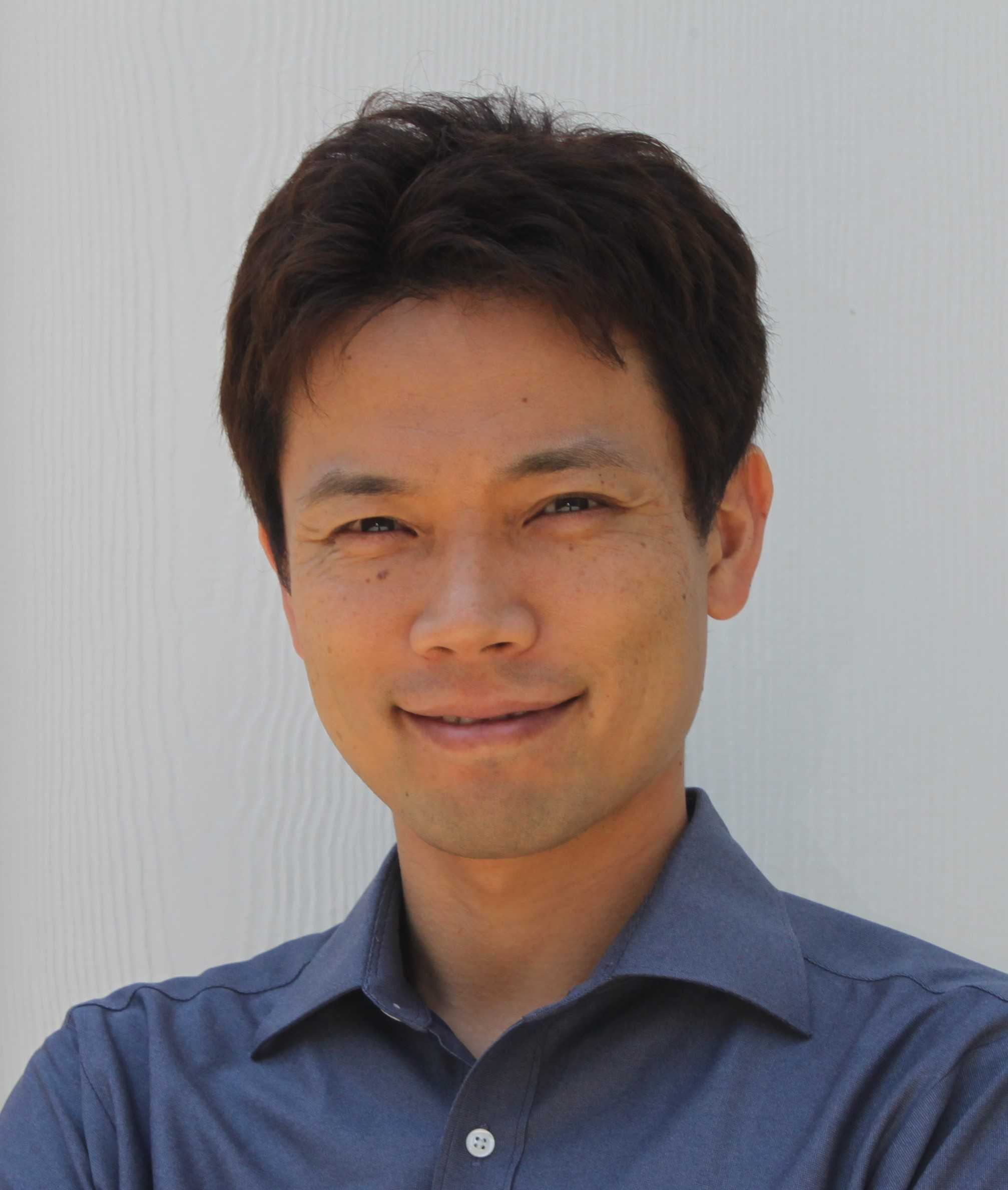}}]{Takashi Tanaka}
Takashi Tanaka received his B.S. (2006) from the University of Tokyo and his M.S. (2009) and Ph.D. (2012) degrees from the University of Illinois at Urbana-Champaign. He was a postdoctoral researcher at the Massachusetts Institute of Technology and KTH Royal Institute of Technology. He was an Assistant Professor at the University of Texas at Austin between 2017 and 2024. Currently, he is an Associate Professor of Aeronautics \& Astronautics and Electrical \& Computer Engineering at Purdue University.
\end{IEEEbiography}

\end{document}